\documentclass[letterpaper]{article} 
\usepackage{aaai2026}  
\usepackage{times}  
\usepackage{helvet}  
\usepackage{courier}  
\usepackage[hyphens]{url}  
\usepackage{graphicx} 
\usepackage{natbib}  
\usepackage{caption} 
\usepackage{algorithm}
\usepackage{algorithmic}
\usepackage{amsmath}
\usepackage{amssymb}
\usepackage{newfloat}
\usepackage{listings}
\DeclareCaptionStyle{ruled}{labelfont=normalfont,labelsep=colon,strut=off} 
\floatstyle{ruled}
\newfloat{listing}{tb}{lst}{}
\floatname{listing}{Listing}
\newtheorem{theorem}{Theorem}[section]
\newtheorem{proposition}[theorem]{Proposition}

\newcommand\modelname{InTRO}

\usepackage{booktabs}
\usepackage[table]{xcolor}
\usepackage{multirow}
\usepackage{subcaption}
\definecolor{Highlight}{HTML}{39b54a}  

\definecolor{Gray}{gray}{0.9}
\usepackage[]{mdframed}
\usepackage{longtable}

\title{In-Token Rationality Optimization: Towards Accurate and Concise LLM Reasoning via Self-Feedback}
\author{%
	Mingye Zhu$^{1,2}$ \quad Yi Liu$^{2}$\thanks{~~Corresponding author: Yi Liu} \quad Zheren Fu$^{1}$\\
	\bf{Quan Wang$^{3}$ \quad Yongdong Zhang$^{1}$} 
} 
\affiliations{
	$^{1}$University of Science and Technology of China, Hefei, China \\
$^{2}$State Key Laboratory of Communication Content Cognition, People's Daily Online, Beijing, China  \\
$^{3}$Beijing University of Posts and Telecommunications, Beijing, China 
}

\usepackage{bibentry}

\begin{document}

\maketitle

\begin{abstract}
	Training Large Language Models (LLMs) for chain-of-thought reasoning presents a significant challenge: supervised fine-tuning on a single “golden” rationale hurts generalization as it penalizes equally valid alternatives, whereas reinforcement learning with verifiable rewards struggles with credit assignment and prohibitive computational cost. To tackle these limitations, we introduce \modelname{} (\textbf{I}n-\textbf{T}oken \textbf{R}ationality \textbf{O}ptimization), a new framework that enables both token-level exploration and self-feedback for accurate and concise reasoning. Instead of directly optimizing an intractable objective over all valid reasoning paths, \modelname{} leverages \emph{correction factors}---token-wise importance weights estimated by the information discrepancy between the generative policy and its answer-conditioned counterpart, for informative next-token selection. This approach allows the model to perform token-level exploration and receive self-generated feedback within a single forward pass, ultimately encouraging accurate and concise rationales.
	Across six math-reasoning benchmarks, \modelname{} consistently outperforms other baselines, raising solution accuracy by up to 20\% relative to the base model. Its chains of thought are also notably more concise, exhibiting reduced verbosity.
	Beyond this, \modelname{} enables cross-domain transfer, successfully adapting to out-of-domain reasoning tasks that extend beyond the realm of mathematics, demonstrating robust generalization.
\end{abstract}

\section{Introduction}

The remarkable success of Large Language Models (LLMs) is highlighted by their emergent ability to tackle complex reasoning and mathematical tasks. A critical breakthrough enabling these capabilities is Chain-of-Thought (CoT) reasoning~\citep{wei2022chain,yu2023metamath,wang2023math,hao2024training}, where models are instructed to generate step-by-step rationales before arriving at a final answer. The conventional approach to teaching CoT, supervised fine-tuning (SFT) with golden rationales, faces crucial limitations: obtaining high-quality step-by-step human annotations is prohibitively expensive, and reliance on single-solution imitation often results in poor generalization as it penalizes equally valid alternative reasoning paths~\citep{kumar2024training,chen2025sft,ni2022learning}.

\begin{figure}[t]
	\centering
	\includegraphics[width=0.98\linewidth]{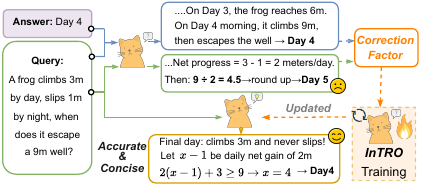} 
	\caption{When solving a reasoning task, the initial model gives a rationale (green) different from its answer-conditioned counterpart (blue), \modelname{} leverages this information discrepancy to compute the correction factors during \modelname{} training, yielding an updated model that produces concise and accurate rationales (orange).}
	\label{fig:exp}
\end{figure}

To enhance reasoning capabilities, recent research has predominantly focused on two alternative paradigms, each with distinct drawbacks. First, reinforcement learning (RL) with verifiable rewards, adopted by models such as OpenAI-o1~\citep{jaech2024openai}, DeepSeek-R1~\citep{guo2025deepseek}, and Kimi-1.5~\citep{team2025kimi}, allows models to bootstrap multiple rationales beyond a single reference answer. However, such RL approaches suffer from sparse, sequence-level feedback that arrives upon rationale completion. This coarse-grained exploration faces the curse of dimensionality---with the space of valid reasoning sequences growing exponentially with length~\citep{gao2025margeimprovingmathreasoning}, further complicating effective credit assignment.

Second, methods incorporating fine-grained feedback or process supervision address reward sparsity by employing \emph{external} ``verifier'' models or human annotators to evaluate individual reasoning steps~\citep{li2023making,weng2022large,chowdhury2025zero,xie2024montecarlotreesearch,zhang2025lessons}. While these process-level signals enhance credit assignment, they introduce challenges including limited labeled data, high annotation costs, significant computational overhead, and potential noise in verifier training.

The aforementioned approaches rely on either coarse-grained exploration or external feedback. Motivated by these limitations, we pose a fundamental question:\textit{
Can a model autonomously explore at the token level while self-generating feedback signals, removing the dependency on external supervision?}
We answer this question affirmatively. However, realizing this capability hinges on solving the intractable objective of \emph{optimal reasoning}.
That is, the model must explore and evaluate reasoning steps in a way that genuinely reflects the true distribution over valid reasoning paths.

To this end, we propose \textbf{I}n-\textbf{T}oken \textbf{R}ationality \textbf{O}ptimization (\modelname{}), a training framework that enables both token-level exploration and self-feedback for accurate and concise reasoning. 
\modelname{} approximates the intractable objective of optimal reasoning by aligning the model’s generative policy $\pi_\theta(z \mid x)$ with its own posterior $\pi_\theta(z \mid x, y)$ via KL divergence minimization, as this alignment yields equivalent gradient updates with optimal reasoning under mild assumptions.
To implement this, we construct an \emph{estimated} posterior by conditioning the model on the correct answer, which we denote as $\pi_\theta(z \mid x\oplus y)$.
This estimated posterior is used to compute \emph{correction factors}—token-wise importance weights that reflect the information discrepancy between the generative policy and the posterior. These correction factors function as self-generated feedback, enabling the model to evaluate how much each token contributes toward the final answer in a single forward pass, thereby facilitating fine-grained, token-level exploration during training.

\begin{figure*}[t]
	\centering
	\includegraphics[width=0.85\linewidth]{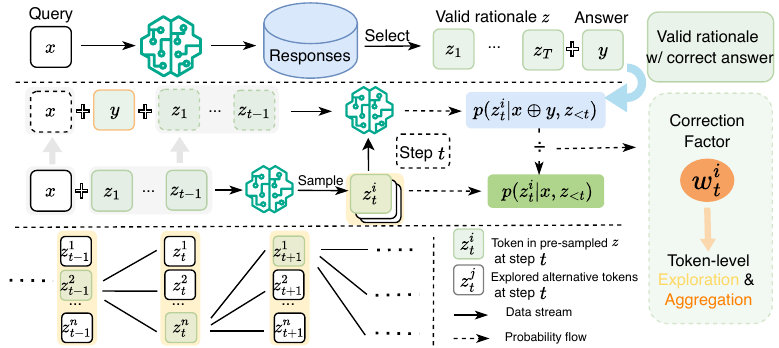} 
	\caption{The illustration of the \modelname{} framework. \emph{Top.} The policy $\pi_\theta$ generates reasoning paths for query $x$ and only paths that yield the correct answer are retained.  
		\emph{Middle.}  For each retained prefix $z_{<t}$ we (i) sample $n$ next tokens $z_{t}^i$ from the forward policy (green) and (ii) obtain the corresponding token probabilities from the estimated posterior by conditioning on the concatenated input $x \oplus y$ (blue).  The ratio of these probabilities gives the token-level correction factor $w_{t}^i$ (orange) for $z_t^{i}$.  
		\emph{Bottom.}  At every position, gradients are aggregated according to $w_{t}^{i}$, providing token-level feedback that guides training.}
	\label{fig:framework}
\end{figure*}

To elucidate \modelname{}'s intuitive operation, we present a simple illustration in Fig.~\ref{fig:exp} and outline the detailed training framework in Fig.~\ref{fig:framework}.
Starting from a query \(x\), the policy \(\pi_\theta\) generates multiple rationales, retaining paths whose answers match the ground truth. For every retained prefix, the model samples \(n\) alternative next-token candidates from its forward policy, obtains the corresponding token probabilities from the estimated posterior by conditioning on the ground truth answer, and computes token-level correction factors. Subsequently, the weighted gradients reinforce the most accurate and logically salient actions, delivering dense, self-generated feedback at every reasoning step.

Empirically, across mathematical reasoning tasks, \modelname{} produces rationales that are remarkably shorter than strong RL baselines while lifting accuracy by up to 20\% relative to the base model. Importantly,
\modelname{} enables cross-domain transfer, successfully adapting to out-of-domain reasoning tasks that extend beyond mathematics, demonstrating robust generalization.

In summary, in this paper we:
\begin{itemize}
	\item propose \modelname{}, a novel token-level exploration and endogenous dense-feedback paradigm for enhancing LLM reasoning capabilities
	\item derive a theoretically grounded, practically feasible learning algorithm that aligns generative and answer-conditioned policies via KL divergence minimization, which equates to solving an intractable optimal reasoning problem under mild conditions
	\item demonstrate that \modelname{} exhibits superior performance and more concise rationales compared to state-of-the-art reasoning methods
	\item provide evidence of enhanced cross-task generalization capabilities, indicating robust and transferable reasoning competence of \modelname{}.
\end{itemize}

\section{In-Token Rationality Optimization}

\subsection{Preliminaries: Goal of Optimal Reasoning}
Let $x$ denote the question, $y$ denote the final answer, and $z=(z_1, z_2, \cdots, z_T)$ represent the CoT reasoning path, where $z_t$ is an intermediate token at step $t$. We use $\pi_\theta$ to denote the model's policy parameterized by $\theta$.
The standard training objective for CoT reasoning via SFT is to maximize the log-likelihood of a single ``golden'' reasoning path $z^*$:
\begin{equation}
	\mathcal{L}_{\text{SFT}} = \mathbb{E}_{(x,z^*,y) \sim \mathcal{D}} \left[-\log \pi_\theta(z^*, y \mid x)\right]
\end{equation}
This approach is fundamentally limited as it forces the model to imitate a single reference solution, thereby assigning low probabilities to alternative but equally valid reasoning paths. A more principled objective is to maximize the marginal log-likelihood of the correct answer $y$~\citep{hoffman2023training}, which involves summing over all valid rationales:
\begin{equation}
	\label{eq:marginal}
	\begin{aligned}
			&\mathcal{L}_{\text{marg.}} = \mathbb{E}_{(x,y) \sim \mathcal{D}} \left[-\log \pi_\theta(y \mid x)\right] \\&= 
			\mathbb{E}_{(x,y) \sim \mathcal{D}} \left[-\log \sum_{z \in \mathcal{Z}_y} \pi_\theta(z,y \mid x)\right],
	\end{aligned}
\end{equation}
where $\mathcal{Z}_y = \{z : f(z) = y\}$. However, directly optimizing this marginal likelihood is intractable due to the exponential growth of the reasoning space.

\subsection{From Intractability to Alignment with Model Posterior}
Given that directly optimizing the marginal likelihood is intractable, we shift our perspective. Rather than explicitly summing over all valid rationales, we ask: \textit{What characterizes optimal reasoning?} Intuitively, an optimal reasoning policy is one whose generative distribution $\pi_\theta(z \mid x)$ naturally emphasizes correct and logically consistent rationales. Such rationales are exactly those the model itself would generate if it already knew the correct answer. Thus, we introduce the model's answer-conditioned posterior, $\pi_\theta(z \mid x,y)$, as an idealized ``teacher'' distribution that embodies correct reasoning.

The challenge then becomes: how do we train the ``student'' policy $\pi_\theta(z \mid x)$ to emulate the ``teacher'' posterior $\pi_\theta(z \mid x,y)$? The most principled way to make one probability distribution resemble another is to minimize the KL divergence between them. In our case, we choose the forward KL divergence as it encourages the policy to broaden its support and better capture diverse valid solutions. This aligns with our goal of improving exploration while still focusing on logically grounded reasoning:
\begin{equation}
\label{eq:kl_objective}
\begin{aligned}
&\min_\theta D_{\text{KL}}(\pi_\theta(z \mid x,y) \parallel \pi_\theta(z \mid x))\\&= \min_\theta \mathbb{E}_{z \sim \pi_\theta(z \mid x,y)} \left[ \log \frac{\pi_\theta(z \mid x,y)}{\pi_\theta(z\mid x)} \right]\\
&= \min_\theta -\mathbb{E}_{z \sim \pi_\theta(z \mid x,y)}[\log \pi_\theta(z \mid x)] + \text{const}.
\end{aligned}
\end{equation}
We treat the posterior as a fixed proposal distribution for estimating the expectation in the KL divergence. Under this view, the entropy term $\mathbb{E}_{z \sim \pi_\theta(z \mid x,y)}[\log \pi_\theta(z \mid x,y)]$ can be approximated as a constant. Therefore, we exclude it from the optimization objective in Eq.~(\ref{eq:kl_objective}).

\subsection{Gradient Equivalence to Marginal Likelihood
}
This KL-based alignment is not merely intuitive, rather, it theoretically equates our original intractable objective of marginal likelihood optimization in Eq.~(\ref{eq:marginal}). 
 Importantly, we first assume that $y$ is a deterministic function of $z$, i.e., $y = f(z)$, where $f$ extracts the final answer from the reasoning path. Next, we establish this relationship through the following proposition (proof provided in Appendix \ref{app:dev}):
\begin{proposition}
	\label{prop:gradient}
	Under the assumption that $y = f(z)$ is a deterministic function of $z$, the gradient of the marginal log-likelihood objective (Eq.~(\ref{eq:marginal})) is identical to the gradient derived from minimizing the KL-divergence objective (Eq.~(\ref{eq:kl_objective})):
	\begin{equation}
		\underbrace{\nabla_\theta \log \pi_\theta(y \mid x)}_{\text{MLL Grad}} = \underbrace{\mathbb{E}_{z \sim \pi_\theta(z \mid x, y)} \left[ \nabla_\theta \log \pi_\theta(z \mid x) \right]}_{\text{Grad derived from KL-minimization}}
	\end{equation}
\end{proposition}
Proposition~\ref{prop:gradient} illustrates that by training the model to approximate the posterior \(\pi_\theta(z \mid x,y)\) over rationales \(z\), we are, in fact, performing a gradient ascent on the marginal log-likelihood of producing the correct answer $y$.

\subsection{Approximation with the Estimated Posterior}
While theoretically appealing, optimizing according to Eq.~(\ref{eq:kl_objective}) involves direct sampling from $\pi_\theta(z\mid x,y)$, which is still infeasible. Therefore, we propose a practical \emph{estimated posterior}, denoted as $\pi_\theta(\cdot\mid x\!\oplus\! y)$, where \(\oplus\) is the concatenation operation. The underlying logic behind this choice leverages modern LLMs' powerful in-context reasoning capabilities.
Empirical evidence demonstrates that LLMs robustly interpret instructions embedded within their inputs~\citep{wei2021finetuned,ouyang2022training,shinn2023reflexion}. Conditioning on both the original question \(x\) and the known correct answer \(y\) simultaneously provides the model with a clear, explicit instructional signal.
Formally, conditioning on $x\!\oplus\! y$ can be viewed as instructing the model to justify a known solution.

 Leveraging this insight, we invoke importance sampling with the proposal $\pi_\theta(z\mid x)$, then Eq.~(\ref{eq:kl_objective}) leads to: 
\begin{equation}
	\mathbb{E}_{z \sim \pi_\theta(z \mid x )} \left[ \frac{\pi_\theta(z \mid x \oplus y )}{\pi_\theta(z\mid x)} \log \pi_\theta(z\mid x) \right].
\end{equation}
Critically, breaking down the above sequence-level importance weight at the token level yields:
\begin{equation}
	w(z) = \prod_{t=1}^{T} \frac{\pi_\theta(z_t \mid x\oplus y,z_{<t})}{\pi_\theta(z_t \mid x, z_{<t})}.
\end{equation}
This allows us to compute token-level correction factors and derive the following practical objective of \modelname{}:
\begin{equation}
	\label{eq:final_loss}
	\mathbb{E}_{(x, y) \sim \mathcal{D}}\left[\frac{1}{|z| \cdot n}\sum_{t=1}^{|z|} \sum_{i=1}^{n} w_{t}^i \cdot \log \pi_\theta(z_{t}^{i} \mid x, z_{<t}) \right] ,
\end{equation}
where
$
w_{t,i}
=\dfrac{\pi_\theta(z_{t}^i\mid x\!\oplus\!y,z_{<t})}
{\pi_\theta(z_{t}^i\mid x,z_{<t})}$ is the \emph{correction score} for token $z_t^i$, and $|z|$ refers to the length of the rational $z$.
Sampling $n$ alternatives per position directly encourages exploration and yields a low-variance gradient
estimate. Tokens with correction factors $w_{t}^i\!>\!1$ reflect high posterior confidence and are reinforced, whereas tokens with
$w_{t}^{i}\!<\!1$ are suppressed. This objective guides generation towards teacher-preferred tokens while refining the teacher’s judgment simultaneously. Moreover, in practice we clip $w_{t}^i \in [0,200]$ to ensure stability during training and curb unreliable teacher effects. The detailed algorithmic steps of \modelname{} are summarized in Algorithm~\ref{alg:japo}.

\paragraph{Difference with traditional RL methods.}
Our \modelname{} objective, as shown in Eq.~(\ref{eq:final_loss}), maximizes the probability of explored actions weighted by the computed correction factors. This fundamentally \emph{differs} from traditional RL algorithms that maximize expected rewards $\mathbb{E}[R]$. Besides, RL methods typically operate by exploring at the sequence level, relying on sparse, rule-based rewards. In contrast, JAPO additionally leverages dense, token-level exploration with self-generated feedback, fostering a more accurate and concise internalization of principled reasoning paths. 

\begin{algorithm}[tb]
	\caption{\modelname{}:In-Token Rationality Optimization}
	\label{alg:japo}
	\textbf{Input}: Policy network \(\pi_\theta\), training data \(\mathcal{D}\), number of sequences to generate per query \(G\), number of tokens to explore per timestep \(n\) \\
	\textbf{Output}: Optimized policy \(\pi^*_\theta\)
	\begin{algorithmic}[1]
		\STATE Initialize \(\theta \leftarrow \theta_{\mathrm{old}}\)
		\FOR{each training iteration}
		\STATE Sample a batch of prompts \(x \sim \mathcal{D}\)
		\FOR{each prompt \(x\)}
		\STATE Generate \(G\) rationales \(\{z_i\}_{i=1}^G \sim \pi_{\theta_{\mathrm{old}}}(\cdot \mid x)\)
		\STATE Filter out invalid rationales (e.g., incorrect final answers)
		\FOR{each valid rationale \(z\)}
		\FOR{each timestep \(t = 1\) to \(T\)}
		\STATE Sample \(n\) token candidates \(\{z_{t}^{j}\}_{j=1}^n\) including the ground-truth token \(z_{t}^{i}\)
		\ENDFOR
		\ENDFOR
			\ENDFOR
		\STATE Compute the surrogate loss \(\mathcal{L}(\theta)\) (see Eq.~(\ref{eq:final_loss}))
		\STATE Update policy: \(\theta \leftarrow \theta + \alpha \nabla_\theta \mathcal{L}(\theta)\)
		\STATE Set \(\theta_{\mathrm{old}} \leftarrow \theta\)
		\ENDFOR
		\STATE \textbf{return} \(\pi^*_\theta\)
	\end{algorithmic}
\end{algorithm}

\section{Experiments}
\subsection{Models and baselines}
We mainly employ the Qwen series: Qwen2.5-1.5B/3B/7B base models and Qwen3-4B/8B base models, as this represents an established practice in the field for their advanced mathematical reasoning capabilities~\citep{wang2025beyond,yan2025learning}.
We compare \modelname{} with several closely-related baselines:
 SFT, {LaTRO}~\citep{chen2024language}, {RAFT++}~\citep{xiong2025minimalist}, {GPG}~\citep{chu2025gpg} and {GRPO}~\citep{shao2024deepseekmath}. Detailed descriptions and objectives for each baseline method are available in Appendix \ref{app:baseline}.
\subsection{Implementation details}
Our implementation builds upon the OpenRLHF framework~\citep{hu2024openrlhf}\footnote{https://github.com/OpenRLHF/OpenRLHF} with 80GB A100 GPUs. For training, we utilize problems with difficulty levels 3-5 from the MATH dataset~\citep{hendrycksmath2021}, comprising approximately 9.2k samples. 
All training experiments use a batch size of 128 and a learning rate of 5e-7. Following the findings of \citet{liu2025understanding}, we avoid prompt templating for the Qwen family during both training and evaluation for better performance.
For each training instance, we generate four candidate responses and assign binary outcome-based rewards (1.0 for correct, 0.0 otherwise). At each step we sample 5 alternative next-token candidates (including the ground-truth token from the original rationale).
To ensure fair comparisons, baseline models are trained using their publicly available codebases and recommended hyperparameter settings, except for RAFT++, which was directly integrated into our experimental setup. Task-specific hyperparameters, including the number of sampled rationales per prompt and the binary reward structure, are held constant across all experiments.

\subsection{Evaluation}

\begin{table*}[t!]
	\centering
	\setlength{\tabcolsep}{0.6mm}
	\begin{tabular}{ccccccccc}
		\toprule
		\textbf{Math Reasoning} & \textbf{MATH500} & \textbf{Minerva Math } & \textbf{Olympiad} & \textbf{College Math}&\textbf{AMC23} & \textbf{AIME25} &\textbf{Average}&\textbf{Impr.(\%)}\\
		\cmidrule{1-9} 	{Accuracy \%}       & pass@1 & pass@1 & pass@1 & pass@1 & avg@32 & avg@32 &&(over base)\\
		\midrule
		Qwen2.5-1.5B & 50.4 & \textbf{11.4}&  14.1&36.7&23.2 &0.6  & 22.7 &-\\
		\multicolumn{1}{c}{ SFT} & 39.2& 7.7  &15.3&34.2 &14.6  &0.5 & 18.6&{-}{18.1}\\
		\multicolumn{1}{c}{ LaTRO} & 46.2 &8.5 &16.1 &36.0  &20.9 & 1.0& 21.5&{-}{5.3} \\
		\multicolumn{1}{c}{ RAFT++} & \underline{53.6}& 8.5&17.3 &36.9  &23.0 &\underline{1.1}&23.4&{+}{3.1}\\
		\multicolumn{1}{c}{ GPG} & 51.2&8.5&16.4&\underline{37.1}&23.8&0.8&23.0&{+}{1.3} \\
		\multicolumn{1}{c}{ GRPO} & 50.6& \underline{9.6}&\underline{17.9}  &37.0& \underline{24.5} &\underline{1.1}&\underline{23.5}&{+}{3.5}\\
	\rowcolor{Gray}	\modelname{} & \textbf{54.2}&\underline{9.6}&\textbf{19.6}&\textbf{38.4}&\textbf{25.5}&\textbf{1.8}& \textbf{24.9}&{+}{9.7}\\
		\midrule
		Qwen2.5-3B & 57.2& \underline{15.4} & 21.5& 38.9 &30.9 &1.6  & 27.6  &-\\
		\multicolumn{1}{c}{ SFT} &44.0 &11.0  &18.1 & 36.4&  19.1& 1.5& 21.7&{-}{21.4}\\
		\multicolumn{1}{c}{ LaTRO} &53.2  &14.3 &21.2 &38.4  & 31.4& 1.4& 26.7&{-}{3.3}\\
		\multicolumn{1}{c}{ RAFT++} & 59.0&15.1 &\textbf{26.4} &40.5  &\underline{35.6} &\textbf{2.8}&29.9&{+}{8.3}\\
		\multicolumn{1}{c}{ GPG} & 57.6&\underline{15.4}&23.3&40.7&34.4&1.9& 28.9&{+}{4.7}\\
		\multicolumn{1}{c}{ GRPO} &\textbf{63.2} &15.1 &\textbf{26.4}  &\underline{41.0} &35.0  & 2.0& \underline{30.5}&{+}{10.5}\\
\rowcolor{Gray}	\modelname{} &\underline{62.6} &\textbf{16.4}&\underline{25.9}&\textbf{41.1}&\textbf{40.1}&\underline{2.3}& \textbf{31.4}&{+}{13.8}\\
		\midrule
		Qwen2.5-7B  & 66.4& 15.1 &30.4 &  42.4&  41.9&5.0  &33.5 & -\\
		\multicolumn{1}{c}{ SFT} & 51.0 & 11.0&23.0 &37.3&  21.6&2.8 &24.5&{-}{26.9} \\
		\multicolumn{1}{c}{ LaTRO} &  65.0&12.5 &29.5 & 42.8 &41.8 & 5.5&32.9&{-}{1.8} \\
		\multicolumn{1}{c}{ RAFT++} & 69.8&14.7 &31.1 &44.0  &44.2 &\textbf{6.6}&35.1&{+}{4.8}\\
		\multicolumn{1}{c}{ GPG} & 69.2&\underline{18.4}&32.1&43.6&44.1&5.3& 35.5&{+}{6.0}\\
		\multicolumn{1}{c}{ GRPO} &\underline{71.8} &17.6 &\underline{33.9}  &\underline{44.7} &\underline{46.2}  &{5.1}& \underline{36.6}&{+}{9.3}\\
	\rowcolor{Gray} \modelname{} & \textbf{72.6}&\textbf{19.9}&\textbf{35.3}&\textbf{45.0}&\textbf{47.0}&\underline{5.6}& \textbf{37.6}&{+}{12.2}\\
		\midrule
		Qwen3-4B & 69.0& 12.5 & 31.3&30.3  & 45.5& \underline{8.9} & 32.9 &-\\
		\multicolumn{1}{c}{ SFT} &65.0& 7.7& 29.9&28.1 & 40.4& 7.3& 29.7&{-}{37.1}\\
		\multicolumn{1}{c}{ LaTRO} &  67.8&12.5 &31.6 &29.6  &47.1  &8.2&32.8&{-}{0.3}  \\
		\multicolumn{1}{c}{ RAFT++} &73.6 & \underline{15.8}&\underline{34.7} & 34.0 & \underline{52.5}&{8.3}&\underline{36.5}&{+}{10.9}\\
		\multicolumn{1}{c}{ GPG} & \underline{74.6}&13.6&34.1&34.1&49.8&7.7&35.7&{+}{8.5}\\
		\multicolumn{1}{c}{ GRPO} & 73.8& \underline{15.8}&  34.4&\underline{34.2}&{50.9}  & 7.9 &36.2&{+}{10.0}\\
	\rowcolor{Gray} \modelname{} & \textbf{74.8}&\textbf{17.6}&\textbf{39.4}&\textbf{35.1}&\textbf{58.3}&\textbf{12.6} &\textbf{39.6}&{+}{20.4}\\
		\midrule
		Qwen3-8B & 65.8&11.8  & 34.7& 29.8 &53.4 & 10.0 & 34.3 &-\\
		\multicolumn{1}{c}{ SFT} & 61.6& 11.0 &34.1 &22.6 &  41.6&8.9 &30.0&-12.5 \\
		\multicolumn{1}{c}{ LaTRO} & 70.0 &12.9 & 33.5&29.3  &52.7 &\underline{11.3} & 35.0&{+}{2.0}\\
		\multicolumn{1}{c}{ RAFT++} & 72.6&12.5 &36.1 &31.1  & 55.4&{10.9}&36.4&{+}{6.1}\\
		\multicolumn{1}{c}{ GPG} &70.4 &\underline{15.4}&33.8&31.4&53.5&10.1& 35.8&{+}{4.4}\\
		\multicolumn{1}{c}{ GRPO} &\underline{74.4} &14.3 & \underline{36.3} &\underline{33.1} &\underline{55.8} & {10.9}&\underline{37.5}&{+}{9.3}\\
	\rowcolor{Gray} \modelname{} & \textbf{75.2}&\textbf{18.8}&\textbf{38.7}&\textbf{35.2}&\textbf{56.7}&\textbf{12.4}&\textbf{39.5}&{+}{15.2} \\
		\bottomrule
	\end{tabular}%
    \caption{Detailed performance of various models across multiple math benchmarks.
		The best result is in bold, the second best one is underscored.
		Our method consistently outperforms baselines across all model scales. Notably, stronger models (e.g., Qwen3 series) show greater performance gains, demonstrating up to 20\% improvement compared to base models.}
	\label{tab:results}
	\end{table*}
Our evaluation focuses on two main categories: In-distribution mathematical reasoning and other out-of-distribution (OOD) tasks.
For mathematical reasoning, we employ a comprehensive suite of benchmarks: MATH500, Minerva Math, OlympiadBench, College Math, AMC23, and AIME25. All evaluations were conducted using the Qwen2.5-Math evaluation codebase~\footnote{https://github.com/QwenLM/Qwen2.5-Math}. We set the sampling temperature to 0.6 and top-p to 0.95.
We report pass@1 for all benchmarks except for AMC23 and AIME25, which have smaller test sets (40 and 30 problems, respectively). For these two tasks, we report avg@32 to mitigate sampling noise, providing a more robust performance indicator compared to pass@1. Consistent with our training procedure, no templates were applied during evaluation. While this might lead to performance disparities compared to templated evaluations (particularly for Minerva Math), it ensures consistency with our training methodology and allows for a clearer assessment of our model's inherent reasoning capabilities. 

To evaluate OOD performance, we test on a diverse set of benchmarks including LiveCodeBench (dynamic programming), BigCodeBench (code understanding and generation), GPQA (open-domain knowledge), HumanEval (function-level code generation), and IFEval (instruction following).

\subsection{Experimental Results}

\paragraph{\modelname{} improves in-distribution mathematical reasoning abilities.}
We conduct a comprehensive comparison with established baselines on several mathematical reasoning tasks, presented in Table~\ref{tab:results}. We see that \modelname{} consistently demonstrates superior performance, frequently achieving the highest accuracy.
Notably, \modelname{} yields more significant performance gains on stronger models, such as the Qwen3 series, indicating the scaling potential of \modelname{}. Furthermore, \modelname{} shows substantial improvements on highly challenging datasets like Olympiad and AIME25, which demand complex, multi-step reasoning. This suggests that token-level exploration and intermediate feedback internalize a more principled and accurate reasoning process, especially for tackling intricate problems. We also provide in Appendix \ref{app:cases} for more case studies.

\begin{table*}[t!]
	\centering
		\begin{tabular}{ccccccc}
			\toprule
			\textbf{Non-math tasks}  & \textbf{LiveCodeBench} & \textbf{BigCodeBench} & \textbf{GPQA}& \textbf{HumanEval}&\textbf{IFEval} &\textbf{Average}\\
			\midrule
			Qwen2.5-7B &7.2 &23.7&36.0&78.7&44.4&38.0 \\
			\multicolumn{1}{c}{ GRPO} & 7.8&\textbf{23.4}&35.0&79.9&48.5&38.9\\
		\rowcolor{Gray}	\multicolumn{1}{c}{ \modelname{}}  & \textbf{11.4}&23.3&\textbf{36.5}&\textbf{81.1}&\textbf{49.8}&\textbf{40.4}\\
			
			\midrule
			Qwen3-4B &4.1 &27.7&22.1&81.7&45.6&36.2  \\
			\multicolumn{1}{c}{ GRPO} & 6.2&27.8&21.9&83.5&40.6&36.0\\
		\rowcolor{Gray}	\multicolumn{1}{c}{ \modelname{}} &\textbf{22.6} &\textbf{35.4}&\textbf{38.6}&\textbf{89.2}&\textbf{50.4}&\textbf{47.2} \\
			\midrule
			Qwen3-8B & 16.0&31.0&36.0&\textbf{86.6}&53.2& 44.6 \\
			\multicolumn{1}{c}{ GRPO} & 17.5&32.3&39.7&\textbf{86.6}&54.0&46.0\\
			\rowcolor{Gray}\multicolumn{1}{c}{ \modelname{}} &\textbf{22.4} &\textbf{35.2}&\textbf{40.6}&{86.0}&\textbf{54.9}&\textbf{47.8}\\
			\bottomrule
		\end{tabular}%
        	\caption{Performance of \modelname{} and GRPO on OOD tasks. Results illustrate that \modelname{} consistently demonstrates improved generalization capabilities compared to GRPO, particularly notable on coding benchmarks for stronger base models.}
	\label{tab:trans_results}
\end{table*}
\paragraph{\modelname{} enhances out-of-distribution generalization performance.} Enabling machines to reason precisely over math is central to automated scientific discovery, but real-world tasks extend far beyond math~\citep{huan2025math}. Therefore, we are curious how do the improved mathematical reasoning abilities transfer to broader capabilities. The results in Tab.~\ref{tab:trans_results} demonstrate that employing \modelname{} extends its benefits beyond pure mathematical tasks, significantly improving performance across a range of OOD tasks compared to GRPO.
These results reveal a critical insight: \modelname{} minimizes token-level information discrepancy to strengthen causal links, enabling logic-driven OOD generalization, especially for coding tasks.

\begin{figure}[htbp]
	\centering
	\begin{subfigure}[b]{0.23\textwidth} 
		\includegraphics[width=\linewidth]{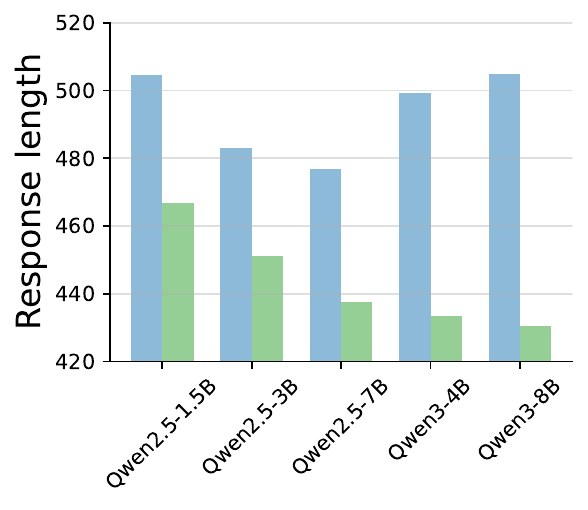}
		\caption{``Hard'' problems.}
	\end{subfigure}
	\hfill
	\begin{subfigure}[b]{0.23\textwidth}
		\includegraphics[width=\linewidth]{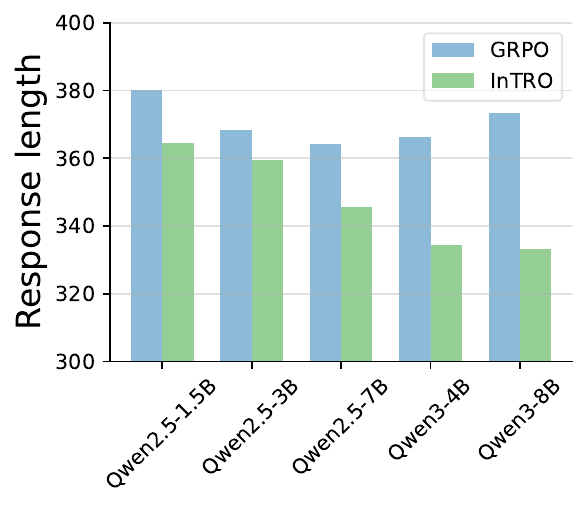}
		\caption{All problems.}
	\end{subfigure}
	\caption{Avg. response length on test questions. \modelname{} provides remarkably shorter rationales, especially on more challenging problems and stronger base models (Qwen3). }
\label{tab:res_length}
\end{figure}

\begin{figure}[htbp]
	\centering
	\begin{subfigure}[b]{0.23\textwidth} 
		\includegraphics[width=\linewidth]{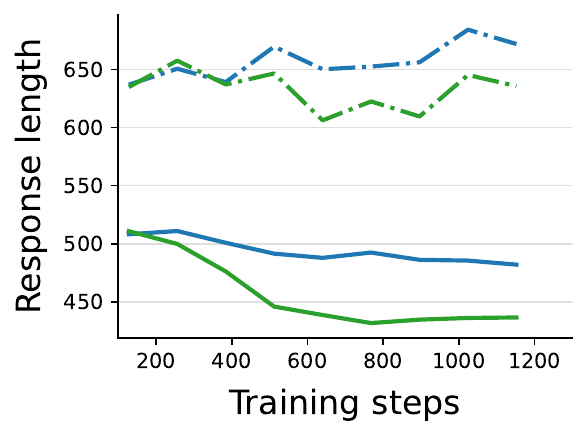}
		\caption{Qwen2.5-7B.}
	\end{subfigure}
	\hfill 
	\begin{subfigure}[b]{0.23\textwidth}
		\includegraphics[width=\linewidth]{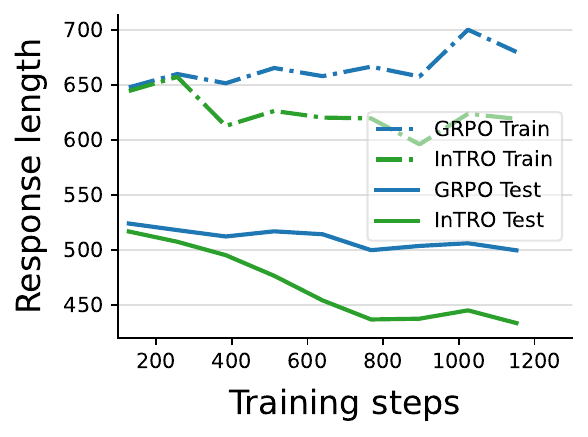}
		\caption{Qwen3-4B.}
	\end{subfigure}
	\caption{Generated response length during training (on training and test questions, respectively).}
	\label{fig:len}
\end{figure}

\begin{figure}[t!]
	\centering
	\begin{subfigure}[b]{0.23\textwidth}
		\includegraphics[width=\linewidth]{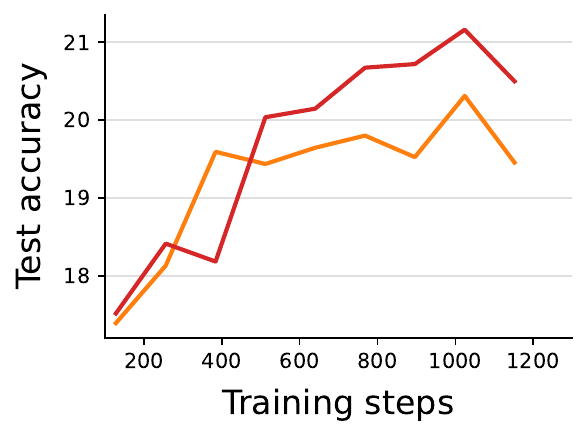}
		\caption{Qwen2.5-7B.}
	\end{subfigure}
	\hfill 
	\begin{subfigure}[b]{0.23\textwidth}
		\includegraphics[width=\linewidth]{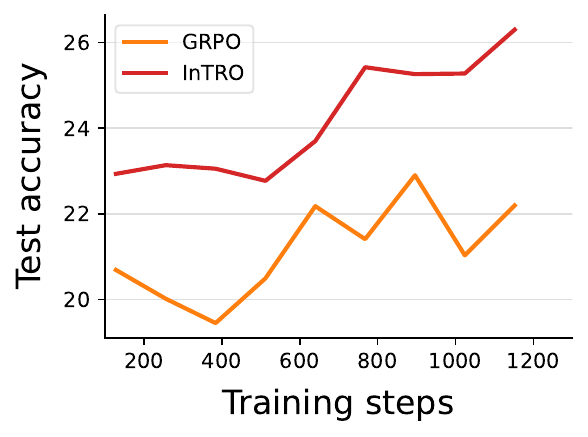}
		\caption{Qwen3-4B.}
	\end{subfigure}
	\caption{Test accuracy on ``Hard'' set  during training.}
	\label{fig:acc}
\end{figure}

\paragraph{\modelname{} optimizes its reasoning trajectories to promote conciseness.}
Fig.~\ref{tab:res_length} reports the average length of the rationales from different test sets, where ``Hard'' denotes challenging problems from Olympiad and AIME25. Training-phase variations in response length (Fig.~\ref{fig:len}) correlate with improved test accuracy on the ``Hard'' set (Fig.~\ref{fig:acc}), revealing \modelname{}’s dual optimization for conciseness and accuracy. This pattern is supported by additional visualizations on training dynamics presented in Appendix \ref{app:dynamics}.\\

\subsection{Ablations}

\begin{table}[htbp]
	\centering
	\setlength{\tabcolsep}{1.2mm}
		\begin{tabular}{ccccccc}
			\toprule
			\# Sampled tokens&1&2&5&10&20&40\\
			\midrule
			Avg. accuracy&20.1&24.4&24.9&25.6&25.0&23.8\\
			\bottomrule
		\end{tabular}%
        	\caption{Effect of different \(n\). Increasing \(n\) improves accuracy by enabling denser token-level exploration, with performance gains gradually saturating beyond a moderate \(n\).}
	\label{tab:diff_n}
\end{table}

\paragraph{Token-level exploration with different \(n\) values.} We investigate with Qwen2.5-1.5B how varying the number of per-step sampled tokens $n$ affects model performance. Specifically, larger $n$ enables more aggressive token-level exploration in \modelname{}. As shown in Tab.~\ref{tab:diff_n}, the accuracy generally improves as we increase $n$, suggesting that more extensive token-level exploration enhances reasoning capabilities. However, beyond a moderate threshold (e.g., \(n=20\)), performance gains plateau then degrade, suggesting that an optimal range for \(n\) exists to maximize model effectiveness. We also provide the 
relationship between larger \(n\) and policy entropy in Appendix \ref{app:entropy} for further analysis.

\begin{figure}[htbp]
	\centering
	
	\begin{subfigure}[b]{0.23\textwidth} 
		\includegraphics[width=\linewidth]{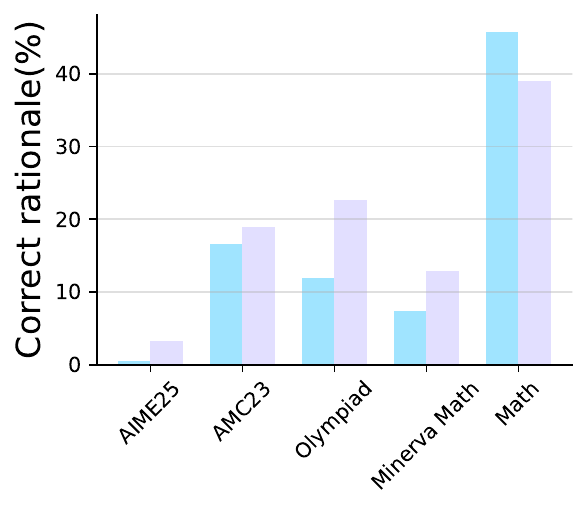}
		\caption{Qwen2.5-1.5B.}
	\end{subfigure}
	\hfill 
	\begin{subfigure}[b]{0.23\textwidth}
		\includegraphics[width=\linewidth]{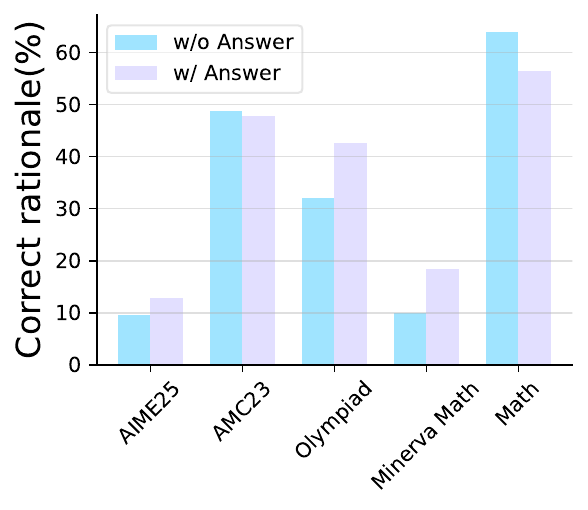}
		\caption{Qwen3-8B.}
	\end{subfigure}
	\caption{Effect of answer-conditioned reasoning. Prompting w/ Answer greatly increases performance on more challenging benchmarks (AIME25 and Olympiad), but yields marginal gains or small drops on easier sets such as Math.}
	\label{fig:backward}
\end{figure}

\paragraph{How well does the answer-conditioned posterior improve reasoning and approximate the true posterior?}
To probe this problem, we ask the model to reason w/ and w/o the final answer, given any question.
To exclude cases where the model simply copy-paste the final answer without giving a genuine rationale, we leverage an external verifier model (Qwen-2.5-70B-Instruct, prompt in Appendix \ref{app:prompt}) to check whether the produced CoT was internally consistent with the answer.
 Fig.~\ref{fig:backward} shows that answer-conditioned reasoning ($\pi_\theta(z | x\oplus y)$) boosts performance on hard sets such as AIME25 and Olympiad, but may slightly hurt on easier ones like Math. The interpretation is intuitive. When the policy faces a vast, high-entropy search space, conditioning on the answer
prunes implausible branches and guides the model toward a valid path. 
For a sanity check, we also estimate the true posterior $\pi_\theta(z|x,y)$ via Bayes’ theorem with $N$ sampled reasoning paths in a toy setup. With sufficiently large $N$(1000 in our experiments), the sequence-level KL divergence between $\pi_\theta(z|x,y)$ and $\pi_\theta(z|x\oplus y)$ is around 2.3, indicating substantial consistency between the two policies.

\paragraph{Computational efficiency analysis.} 
While \modelname{} does token‑level exploration with $n$ candidates at each step, it simply reweights pre‑computed logits, adding virtually no extra cost. As a result, for each rationale, only two forward passes are required to compute the correction factors. In contrast, GRPO must proportionally increase the number of fully sampled trajectories $G$ to stimulate exploration. This requires $2G$ forward passes to evaluate each rationale under both the new and old policies.
Consequently, \modelname{} maintains significantly lower logit-compute overhead compared to GRPO, especially as the exploration density increases.

\section{Related work}
\subsection{Improving the reasoning chains}
CoT reasoning techniques have been shown to be effective in instructing modern LLMs to solve complex reasoning tasks. A large body of works fall into this category, such as prompting methods~\citep{wei2022chain,khot2022decomposed} and tuning-based methods~\citep{yu2023metamath, hao2024training,cheng2024compressed,wang2025critique}. One line of work focuses on improving the correctness of the reasoning steps. For instance, GRACE~\citep{khalifa2023grace} first learn a discriminator over correct and incorrect steps, then leverages it during decoding to score next-step candidates. CFT~\citep{wang2025critique} improves traditional SFT by explicitly training models to critique noisy responses and verify correctness via maximizing the likelihood of the annotated critique. Another line of work explores latent reasoning in LLMs~\citep{pfau2024let,deng2023implicit,deng2024explicit}. Specifically, Coconut~\citep{hao2024training} directly feeds the last hidden state as the input embedding for the next token prediction and enables reasoning in continuous space, and CCoT~\citep{cheng2024compressed} produces contemplation tokens for additional reasoning over dense representations.  Another literature explores self-play~\citep{zelikman2022star,qu2024recursive,kumar2024training,zelikman2024quiet}. TRICE~\citep{hoffman2023training} leverages Markov-chain Monte Carlo expectation-maximization to maximize the marginal log-likelihood of generating a correct answer using CoT prompting, and LaTRO~\citep {chen2024language} formulates reasoning as sampling from a latent distribution and optimizes it via variational approaches, with RLOO employed to optimize the reasoner~\citep{kool2019buy}
Our work, while also trying to improve the rationale quality, lets the model’s \emph{own} posterior serve as the teacher, supplying token-level feedback during training.

\subsection{LLMs for mathematical reasoning}
With the rapid development of the reasoning-centered LLMs, especially following the release of GPT-o1~\citep{jaech2024openai} and DeepSeek-R1~\citep{guo2025deepseek}, the research focus has shifted to leveraging RL-based algorithms to boost reasoning capabilities. Specifically, 
GRPO~\citep{shao2024deepseekmath} is a representative work that greatly simplifies the RL process by eliminating the value function in Proximal Policy Optimization (PPO)~\citep{schulman2017proximal}. It leverages the behavior policy to sample a group of responses and calculate the advantage by normalizing the group-level rewards. GRPO has been shown to be really impressive and inspired subsequent works such as Dr.GRPO~\citep{liu2025understanding}, DAPO~\citep{yu2025dapo}, SEED-GRPO~\citep{chen2025seed}, and EMPO~\citep{zhang2025right}. Rejection sampling fine-tuning, or RAFT~\citep{dong2023raft}, turns out to be a highly comparable alternative to complex RL algorithms~\citep{xiong2025minimalist} by fine-tuning the models on the self-generated correct generations. \citet{xiong2025minimalist} also proposes RAFT++, which is an improved RAFT with importance sampling and gradient clipping, and Reinforce-Rej, which is an RL variant that filters out both fully incorrect and correct samples. Similarly, GPG~\citep{chu2025gpg} directly optimizes the original RL objective, with a thresholding mechanism to avoid large variance in gradient estimation. The proposed \modelname{} sits at the intersection of these lines. It re-uses the model’s own generations, but attaches a soft, answer-conditioned weight to every explored token, providing a new avenue for accurate and concise reasoning.
\section{Conclusion}
In this paper, we introduce \modelname{} to enable accurate and concise reasoning by unifying token-level exploration with self-generated feedback, all without external guidance. \modelname{} training is theoretically grounded as it is gradient-equivalent to maximizing the intractable goal of optimal reasoning. The derived practical implementation involves an estimated posterior to compute correction factors in a single forward pass, ultimately encouraging accurate and concise rationales. Experiments on multiple mathematical and other reasoning tasks show that \modelname{} provides consistent improvement over other baselines to a great margin.

\section*{Acknowledgments}
			This research is supported by Artificial Intelligence-National Science and Technology Major Project 2023ZD0121200.

\bibliography{aaai2026}

\newpage

\appendix

\section{Mathematical derivation}
\label{app:dev}

We aim to show that under the assumption that $y = f(z)$ is a deterministic function of $z$, the gradient of the marginal log-likelihood $\log \pi_\theta(y \mid x)$ is equal to the gradient obtained by minimizing the KL divergence $D_{\mathrm{KL}}(\pi_\theta(z \mid x, y) \, \| \, \pi_\theta(z \mid x))$.

Under the deterministic assumption $y = f(z)$, we can write:
\[
\pi_\theta(y \mid x) = \sum_{z \in \mathcal{Z}_y} \pi_\theta(z \mid x)
\]
Therefore, using the log-derivative trick, where $\nabla f(\theta) = f(\theta) \nabla \log f(\theta)$ (we can think of \(f(\theta)\) first as \(\sum_{z \in \mathcal{Z}_y} \pi_\theta(z \mid x)\) then \(\pi_\theta(z \mid x)\)), the gradient of Eq.(2) with respect to the model parameters $\theta$ can be expressed as:
\begin{align*}
	&\nabla_\theta \log \pi_\theta(y \mid x)= \nabla_\theta \log \sum_{z \in \mathcal{Z}_y} \pi_\theta(z \mid x) \\
	&= \frac{1}{\sum_{z \in \mathcal{Z}_y} \pi_\theta(z \mid x)} \sum_{z \in \mathcal{Z}_y} \nabla_\theta \pi_\theta(z \mid x) \\
	&= \frac{1}{\pi_\theta(y \mid x)} \sum_{z \in \mathcal{Z}_y} \pi_\theta(z \mid x) \nabla_\theta \log \pi_\theta(z \mid x) \\
	&= \sum_{z \in \mathcal{Z}_y} \frac{\pi_\theta(z \mid x)}{\pi_\theta(y \mid x)} \nabla_\theta \log \pi_\theta(z \mid x)\\
	&=\sum_{z \in \mathcal{Z}_y}\pi_\theta(z \mid x,y)\nabla_\theta \log \pi_\theta(z \mid x).
\end{align*}

This is the expectation over the conditional distribution $\pi_\theta(z \mid x, y)$:
\[
\nabla_\theta \log \pi_\theta(y \mid x)
= \mathbb{E}_{z \sim \pi_\theta(z \mid x, y)} \left[ \nabla_\theta \log \pi_\theta(z \mid x) \right]
\]

Consider the KL divergence:
\[
D_{\mathrm{KL}}(\pi_\theta(z \mid x, y) \, \| \, \pi_\theta(z \mid x)) = \mathbb{E}_{z \sim \pi_\theta(z \mid x, y)} \left[ \log \frac{\pi_\theta(z \mid x, y)}{\pi_\theta(z \mid x)} \right]
\]

Taking the gradient with respect to $\theta$, ignoring the constant:
\[
\nabla_\theta D_{\mathrm{KL}} = - \mathbb{E}_{z \sim \pi_\theta(z \mid x, y)} \left[ \nabla_\theta \log \pi_\theta(z \mid x) \right]
\]

Thus, minimizing the KL divergence yields:
\[
\nabla_\theta \log \pi_\theta(y \mid x) = \mathbb{E}_{z \sim \pi_\theta(z \mid x, y)} \left[ \nabla_\theta \log \pi_\theta(z \mid x) \right]
\]
This completes the proof that, under the deterministic mapping $y = f(z)$, the gradient of the marginal log-likelihood is equivalent to the gradient derived from minimizing the KL divergence.

\section{Experiments}
\subsection{Baselines}
\label{app:baseline}
We compare \modelname{} with several recent, closely-related baselines specifically designed to enhance reasoning capabilities through self-generated rationales:

\textbf{LaTRO}~\citep{chen2024language} proposes to leverage $\log p_{\theta}(y | ~x\oplus z)$ as the reward function to evaluate the quality of the rationale $z$ given the pair $(x, y)$, without requiring external feedback or reward models: 
\begin{align*}
	& \mathbb{E}_{(x, y)\sim \mathcal{D}} \bigg[\mathbb{E}_{z \sim \pi_{\theta}(\cdot | x)}\log \pi_{\theta}(y | ~x\oplus z)\big] \\
	& - \text{KL}[\pi_\theta(z |x)|| \pi_0(z | x)]  \bigg]\,. \label{eq:latro}
\end{align*}

\textbf{RAFT++}~\citep{xiong2025minimalist} applies the importance sampling and clipping techniques to the original RAFT, which trains only on positively rewarded samples, yielding competitive performance to GRPO:
\begin{equation*}
	\begin{aligned}
		\mathbb{E}_{(x,y) \sim \mathcal{D}} \frac{1}{|z|}\sum_{t=1}^{|z|} \Big[ \min\Big(r_t(\theta), \text{clip}(r_t(\theta), 1-\epsilon,1+\epsilon)\Big)\Big],
	\end{aligned}
\end{equation*}
where the rationales \(z\) are filtered to only preserve those that leads to the correct answers \(y\).

\textbf{GPG}~\citep{chu2025gpg} directly optimizes the original RL objective and significantly simplifies the training process compared to GRPO:
\begin{equation*}
	\begin{aligned}
		& \mathbb{E}_{(x,y) \sim \mathcal{D}, \{z_i\}_{i=1}^{G}} \\
		&	 \left[	\frac{1}{\sum_{i=1}^{G}{|z_i|}} \sum_{i=1}^{G}  \sum_{t=1}^{|z_i|}  \left( \log \pi_{\theta}(z_{i,t} \mid x, z_{i,<t}) \hat{A}_{i,t} \right)\right] ,
	\end{aligned}
\end{equation*}
where
\begin{equation*}
	\hat{A}_{i,t} = \frac{r_i - \text{mean}(\{R_i\}_{i=1}^{G})}{{F_{norm}}}.
\end{equation*}

\textbf{GRPO}~\citep{shao2024deepseekmath} obviates the need for value function approximation in traditional RL, and uses the average reward of multiple sampled outputs to the same question, as the baseline:
\begin{equation*}
	\begin{aligned}
		&\mathbb{E}_{(x,y)\sim \mathcal{D}, \{z_i\}_{i=1}^G} 
		\Bigg[ \frac{1}{G}\sum_{i=1}^{G} \frac{1}{|z_i|}\sum_{t=1}^{|z_i|} \Bigg( 
		\min \Big( r_{i,t}(\theta),	\\&	
		\ \text{clip} ( r_{i,t}(\theta), 1 - \varepsilon, 1 + \varepsilon )  \Big)\hat{A}_{i,t}
		- \beta D_{\text{KL}}(\pi_{\theta} || \pi_{\text{ref}}) 
		\Bigg) \Bigg],
		\label{eq:grpoloss}
	\end{aligned}
\end{equation*}
where
\begin{equation*}
	r_{i,t}(\theta)=\frac{\pi_{\theta}(z_{i,t} \mid x, z_{i,<t})}{\pi_{\theta_{\text{old}}}(z_{i,t} \mid x,z_{i,<t})},
	\hat{A}_{i,t} = \frac{r_i - \text{mean}(\{R_i\}_{i=1}^G)}{\text{std}(\{R_i\}_{i=1}^G)}.
\end{equation*}

\subsection{More training dynamics}
\label{app:dynamics}
\begin{figure}[htbp]
	\centering
	\begin{subfigure}[b]{0.23\textwidth} 
		\includegraphics[width=\linewidth]{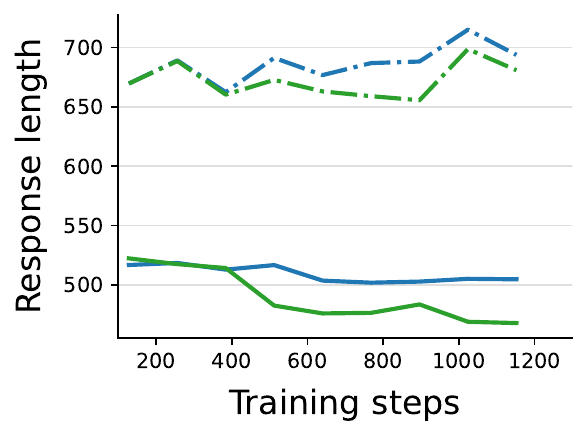}
		\caption{Qwen2.5-1.5B.}
	\end{subfigure}
	\hfill 
	\begin{subfigure}[b]{0.23\textwidth}
		\includegraphics[width=\linewidth]{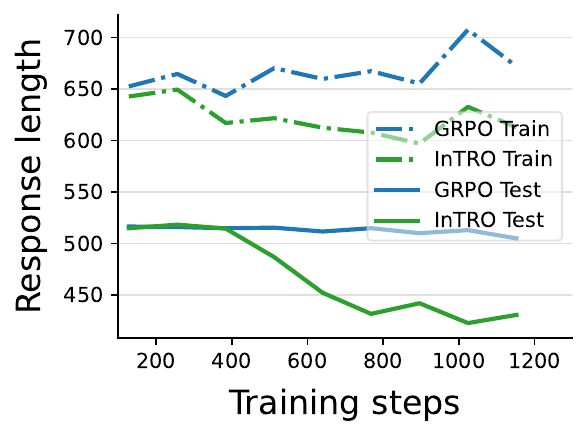}
		\caption{Qwen3-8B.}
	\end{subfigure}
	\caption{Generated response length during training (on training and test questions, respectively).}
	\label{fig:more-len}
\end{figure}

\begin{figure}[t!]
	\centering
	\begin{subfigure}[b]{0.23\textwidth} 
		\includegraphics[width=\linewidth]{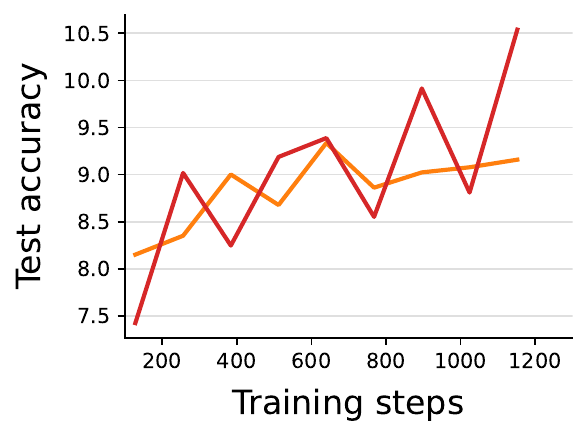}
		\caption{Qwen2.5-1.5B.}
	\end{subfigure}
	\hfill 
	\begin{subfigure}[b]{0.23\textwidth}
		\includegraphics[width=\linewidth]{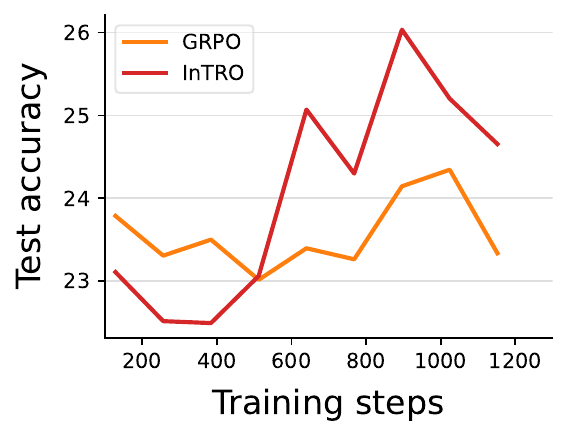}
		\caption{Qwen3-8B.}
	\end{subfigure}
	\caption{Test accuracy on ``Hard'' set during training.}
	\label{fig:more-acc}
\end{figure}

In Figures~\ref{fig:more-len} and~\ref{fig:more-acc}, we present additional length variation statistics during training for Qwen2.5-1.5B and Qwen3-8B, along with their corresponding test accuracy. These statistics are derived from OlympiadBench and AIME25, as we consider these benchmarks more challenging and better suited for evaluating a model’s reasoning capabilities. The observed trend aligns with the findings in the main paper: \modelname{} reduces the average test length while improving accuracy on these tasks.

\begin{figure}[t!]
	\centering
	\begin{subfigure}[b]{0.23\textwidth} 
		\includegraphics[width=\linewidth]{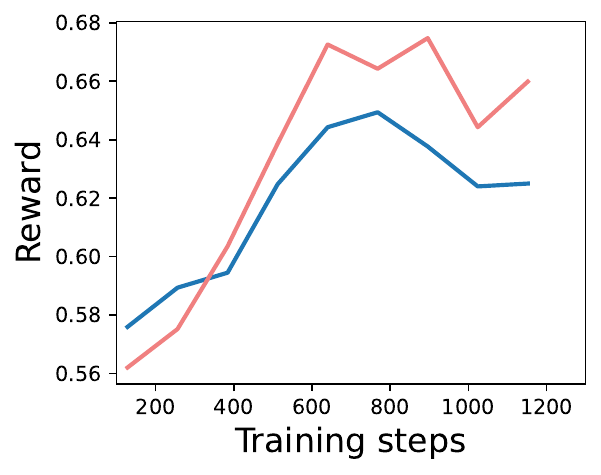}
		\caption{Qwen2.5-7B}
	\end{subfigure}
	\hfill 
	\begin{subfigure}[b]{0.23\textwidth}
		\includegraphics[width=\linewidth]{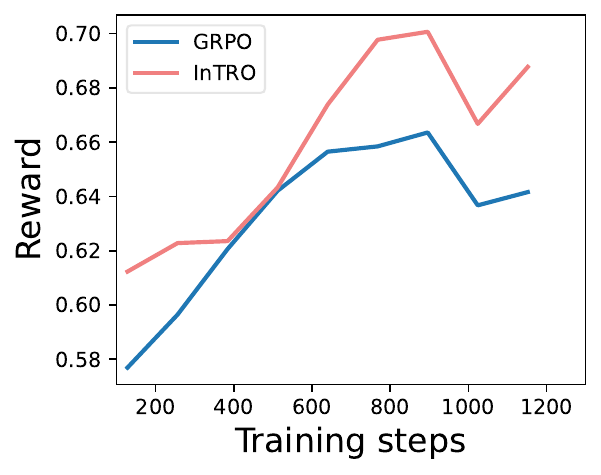}
		\caption{Qwen3-4B.}
	\end{subfigure}
	\caption{Average reward during training.}
	\label{fig:reward}
\end{figure}
We further analyze the training rewards for Qwen2.5-7B and Qwen3-4B base models in Figure~\ref{fig:reward}. Although \modelname{} is not explicitly optimized to maximize reward during training, it consistently achieves higher reward values than GRPO. This reinforces our proposition that the practical token-level implementation effectively aligns with the desired optimization objective.

\subsection{More analysis on answer-conditioned reasoning}

\paragraph{Evaluation prompt for the CoT verifier.}
\label{app:prompt}
When evaluating the effectiveness of answer-conditioned reasoning, we leverage Qwen2.5-72B-Instruct as the verifier to assess if the reasoning path given by the base model is consistent with the final answer. The evaluation prompt is as follows:
\begin{mdframed}[linewidth=1pt, linecolor=black] 
	
	Act as a meticulous verifier. Your task is to determine whether the provided solution correctly and logically leads to the given final answer. Follow these steps:\\
	
	Examine the Solution: Carefully analyze the reasoning, calculations, and steps in the solution.\\
	
	Check for Consistency: Verify that each step follows logically from the previous one and that no critical errors (mathematical, logical, or factual) are present.\\
	
	Validate the Final Answer: Confirm whether the final answer is a direct and correct result of the solution’s steps. Provide a clear judgment:\\
	
	If consistent: State [Consistent] and briefly explain why.\\
	
	If inconsistent: State [Inconsistent] and point out where the error(s) occur and why they invalidate the conclusion.\\
	
	Format your response as follows:\\
	
	Judgment: [Consistent/Inconsistent]\\
	
	Reasoning: [Your analysis]\\
	
	Below is the solution to evaluate:\\
	\texttt{<CoT>}  
	
	Below is the final answer:\\
	\texttt{<Answer>}  
	
\end{mdframed}

\begin{figure}[htbp]
	\centering
	\begin{subfigure}[b]{0.23\textwidth} 
		\includegraphics[width=\linewidth]{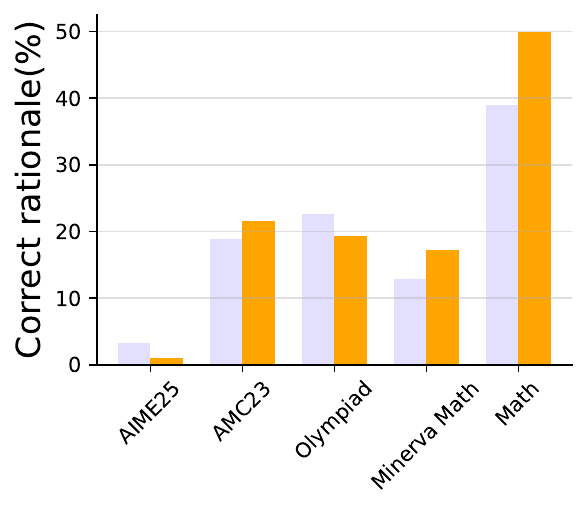}
		\caption{Qwen2.5-1.5B.}
	\end{subfigure}
	\hfill 
	\begin{subfigure}[b]{0.23\textwidth}
		\includegraphics[width=\linewidth]{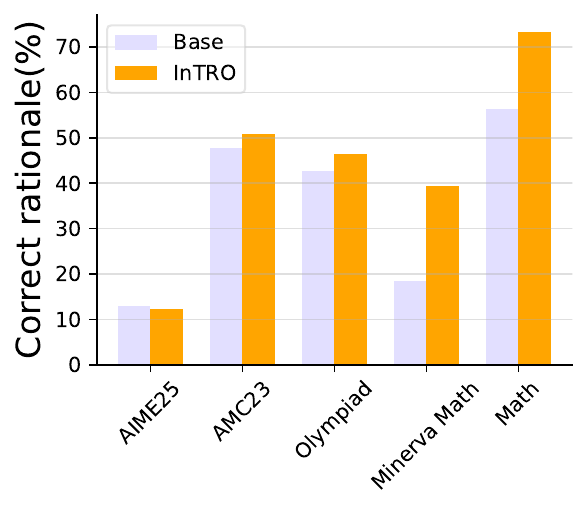}
		\caption{Qwen3-8B.}
	\end{subfigure}
	\caption{Performance of answer-conditioned reasoning with base model and \modelname{}-trained models.\modelname{} training greatly enhances the model's ability to reason w/ answers and enables more correct rationales.}
	\label{fig:backwards}
\end{figure}

\paragraph{\modelname{} training enhances answer-conditioned reasoning.} We evaluate whether models trained with \modelname{} retain their answer-conditioned reasoning capabilities, with comparative results against the base model shown in Figure~\ref{fig:backwards}. Notably, we find that answer-conditioned reasoning performance actually improves post-training with \modelname{}. This enhancement suggests that by aligning the policy with its answer-conditioned posterior, the model develops improved capacity to generate consolidated rationales in a more principled way.

\subsection{Results on models other than Qwen series}
\label{app:llama}
Our experiments primarily focus on the Qwen series models, renowned for their strong reasoning capabilities, particularly in mathematical tasks. To further investigate the generalizability of \modelname{}, we also evaluate its application to base models with weaker inherent reasoning abilities, using Llama-3.1-8B-Instruct and Llama-3.2-3B-Instruct as a test case. Key results are presented in Table~\ref{tab:llama}. Our results demonstrate varying degrees of performance improvement across all methods compared to the base model. As \modelname{} relies on the model's self-generated correction factors derived from the estimated posterior, we observe that weaker models fail to match the performance levels of their stronger counterparts. However, \modelname{} still demonstrates a certain advantage over other baselines tested.

\subsection{Case study}
\label{app:cases}
In this section, we provide two examples from the challenging OlympiadBench with Qwen3-4B base model aligned with GRPO and \modelname{}, respectively. From Table~\ref{tab:cases} we observe that across the two examples \modelname{}’s rationales average 234 words, versus 410 words for GRPO, yet still include all essential algebraic steps. This pattern underscores that our guided-reward training not only improves correctness but also suppresses superfluous reasoning chains.

\onecolumn

\begin{table*}[t!]
	\centering
	
	\setlength{\tabcolsep}{0.6mm}
	\begin{tabular}{ccccccccc}
		\toprule
		\textbf{Math Reasoning} & \textbf{MATH500} & \textbf{Minerva Math } & \textbf{Olympiad} & \textbf{College Math}&\textbf{AMC23} & \textbf{AIME25} &\textbf{Average}&\textbf{Impr.(\%)}\\
		\cmidrule{1-9} 	{Accuracy \%}       & pass@1 & pass@1 & pass@1 & pass@1 & avg@32 & avg@32 &&(over Base)\\
		\midrule
		Llama-3.2-3B-Instruct& 33.4 &10.3&11.0 &11.2&22.7&0.6 &14.9 &-\\
		\multicolumn{1}{c}{ RAFT++} &37.2 &10.3 &12.4& 15.1&22.7&0.6& 16.4&+10.1 \\
		\multicolumn{1}{c}{ GRPO} &40.8 &14.0 &12.3& 15.8&23.9&1.1& 18.0&+20.8 \\
		\modelname{} &38.4&16.5 &12.6&23.5 &24.3&1.0& 19.4 & +30.2\\
		\midrule
		Llama-3.1-8B-Instruct &37.0&17.3&13.0&12.3&21.0&0.9&16.9&-\\
		RAFT++ &45.6&21.0&16.4&26.6&20.3&0.3&21.7&28.4\\
		GRPO &44.4&20.6&16.9&27.6&23.2&0.9&22.3&32.0 \\
		\modelname{} &45.0&22.1&14.1&28.9&23.4&2.0&22.6&33.7\\
		\bottomrule
	\end{tabular}%
\caption{Mathematical reasoning performance with other models. }
\label{tab:llama}
\end{table*}

\begin{longtable}{@{}p{0.5\textwidth}p{0.5\textwidth}@{}}
\toprule\noalign{\smallskip}
	\multicolumn{2}{@{}p{\textwidth}@{}}{\textbf{Query 1:} 	Let $\lfloor x\rfloor$ represent the greatest integer $\le x$.  If $x$ is positive and $x\lfloor x\rfloor = 17$, what is the value of $x$?} \\
	\midrule
	\textbf{Response 1 (GRPO):} & \textbf{Response 2 (\modelname{}):} \\ 
	
	\scriptsize
	To solve the equation \( x \lfloor x \rfloor = 17 \) for a positive \( x \), we need to find a value of \( x \) such that when we multiply \( x \) by its floor value, we get 17.
	
	\textbf{Step 1: Understand the equation.}  
	– \( \lfloor x \rfloor \) is the greatest integer $\le x$.  
	– We seek \( x \) so that \( x\lfloor x \rfloor =17\).
	
	\textbf{Step 2: Set up the equation.}  
	Let $n=\lfloor x\rfloor$.  Then $n$ is an integer and $n\le x<n+1$.  The equation becomes $xn =17$ or $x=17/n$.
	
	\textbf{Step 3: Search integer candidates.}  
	Because $x< n+1$, we require $\dfrac{17}{n} < n+1\;\Longrightarrow\;17 < n(n+1)$.  Also $x\ge n$ gives $\dfrac{17}{n}\ge n\;\Longrightarrow\;17\ge n^{2}$.
	
	Thus $n^{2}\le17< n(n+1)$.  Checking $n=4$ fails because $17<4\cdot5=20$.  For $n=3$ we have $3^{2}=9\le17<3\cdot4=12$ (false).  Continuing\ldots{} the only integer that seems to fit the inequality analysis is $n=5$, but $5^{2}=25>17$, contradiction.
	
	After a python‐assisted check we obtain one numeric solution:
	\begin{verbatim}
		>>> n_candidates = [n for n in range(1,18)
		if n*n <=17 < n*(n+1)]
		>>> n_candidates
		[3]
		>>> x = 17/3
		3.40000000000000
	\end{verbatim}
	Hence the solution is $\boxed{3.4}$.
	&
	\scriptsize
	Let $n=\lfloor x\rfloor$.  Then $n\le x<n+1$ and $xn=17$.
	
	\textbf{1. Bound $n$.} Because $x > n$,  
	\[
	17 = xn \ge n^{2}\;\;\Longrightarrow\;\;n^{2}\le17.
	\]
	Because $x<n+1$, 
	\[
	17 = xn < n(n+1)\;\;\Longrightarrow\;\;17< n(n+1).
	\]
	Thus $n^{2}\le17< n(n+1)$.
	
	\textbf{2. Find the unique integer $n$.}
	Try $n=4$: $4^{2}=16\le17<4\cdot5=20$ holds, so $n=4$.  
	Try $n=3$: $17<3\cdot4=12$ fails.  
	Hence $n=4$ is the only integer.
	
	\textbf{3. Solve for $x$.}\,
	\(x = \dfrac{17}{n} = \dfrac{17}{4} = 4.25.\)
	
	Therefore $x=\boxed{4.25}$.\\ 
	\hline
	\textbf{Answer:} &\\
	{Wrong}\,: $3.4$ \;(455 words) &{Correct}\,: $4.25$ \;(288 words) \\ \bottomrule
	\hline
	
	\multicolumn{2}{@{}p{\textwidth}@{}}{\textbf{Query 2:} Paul planned to buy $20$ souvenirs.  Before paying he removed $40\%$ of them, which lowered the bill by \$108.  How many dollars did he spend on the remaining items?}\\ \midrule
	\textbf{Response 1 (GRPO):} & \textbf{Response 2 (\modelname{}):}\\
	\scriptsize
	First note that removing $40\%$ of 20 items means taking away $0.4\cdot20=8$ items, leaving $12$.
	
	Assume each souvenir costs $p$ dollars.  The original total would have been $20p$.  After removing $8$ items the bill should decrease by $8p$.  We are told this reduction equals \$108, so $8p=108$ and thus $p=13.5$.
	
	At this point, one might think the cost of the remaining $12$ items is $12p=162$.  However, because the problem says “reduced by \$108,” another interpretation is that \$108 \emph{is exactly} what Paul ended up paying.  Choosing the latter (it seems more natural in contest phrasing), the answer is $\boxed{108}$. 
	&
	\scriptsize
	Let each souvenir cost $p$ dollars.
	
	\textbf{1. Relate item count to price.} 
	Removing $40\%$ of $20$ items means removing $8$ items, leaving $12$.
	
	\textbf{2. Translate \$108 reduction.} 
	The price drop equals the cost of those $8$ items: $8p = 108 \Rightarrow p = 9$.
	
	\textbf{3. Compute cost of remaining items.} 
	Remaining bill $= 12p = 12 \times 9 = \boxed{120}$. \\ \midrule
	\textbf{Answer:}&\\
	{Wrong}: \$108\;(365 words) & {Correct}: \$120\;(181 words)\\ \bottomrule
	\label{tab:cases}
\end{longtable}

\twocolumn
\subsection{Visualizations of token entropy}
\label{app:entropy}
\begin{figure}[htbp]
	\centering
	\begin{subfigure}[b]{0.23\textwidth} 
		\includegraphics[width=\linewidth]{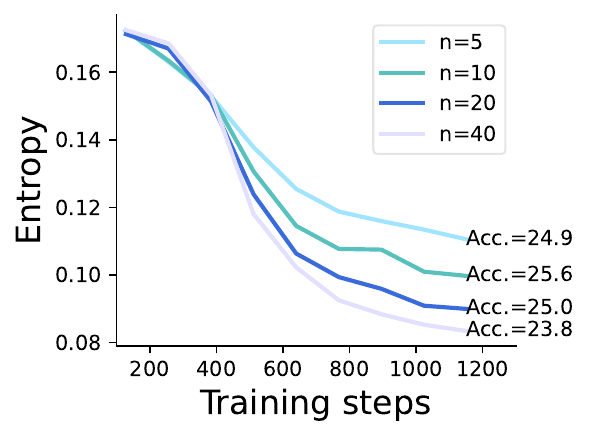}
		\caption{$n$ tokens explored in \modelname{}.}
	\end{subfigure}
	\hfill 
	\begin{subfigure}[b]{0.23\textwidth}
		\includegraphics[width=\linewidth]{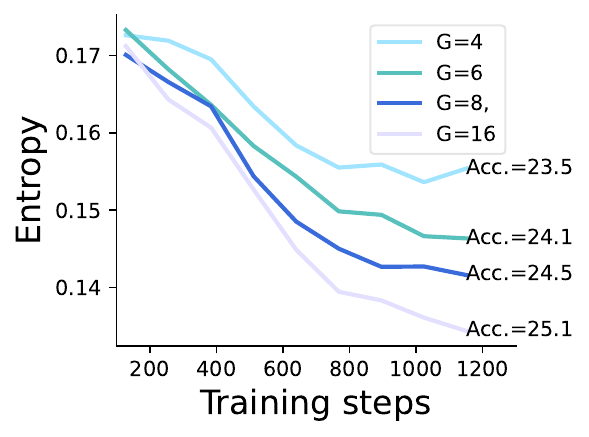}
		\caption{$G$ sequences for GRPO.}
	\end{subfigure}
	\caption{Entropy dynamics and performance of \modelname{} and GRPO under varying exploration granularity.}
	\label{fig:entropy}
\end{figure}

\begin{figure*}
	\includegraphics[width=\linewidth]{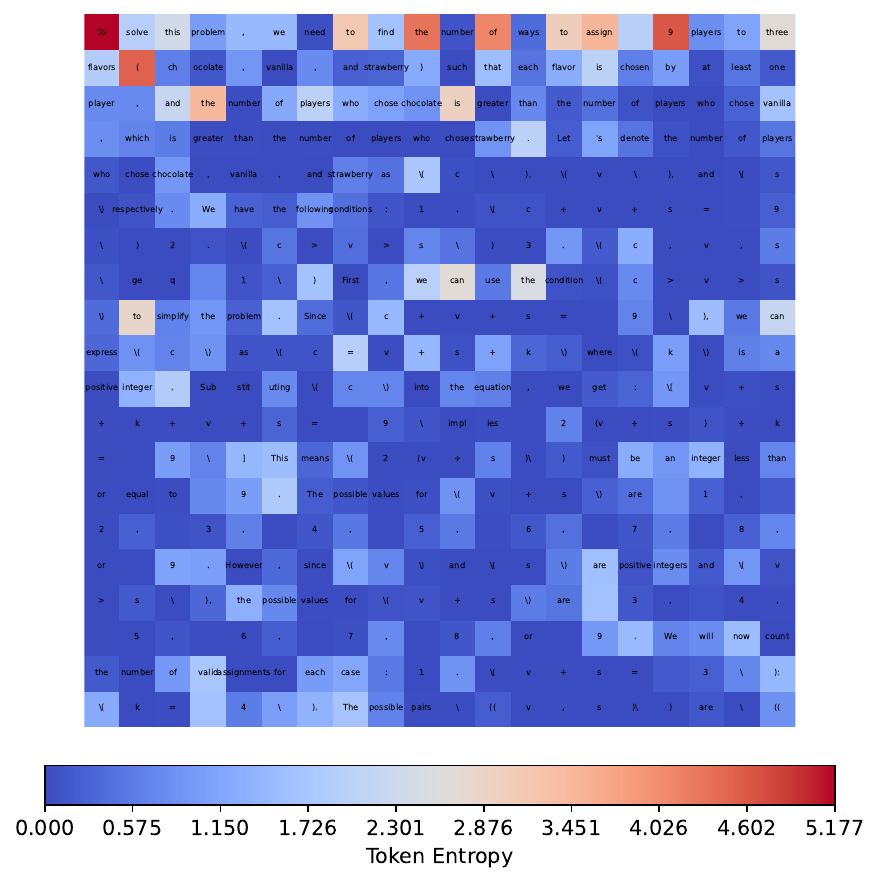}
	\caption{Random example 1 from AIME25 with GRPO}
	\label{fig:entropy-grpo-1}
\end{figure*}

\begin{figure*}
	\includegraphics[width=\linewidth]{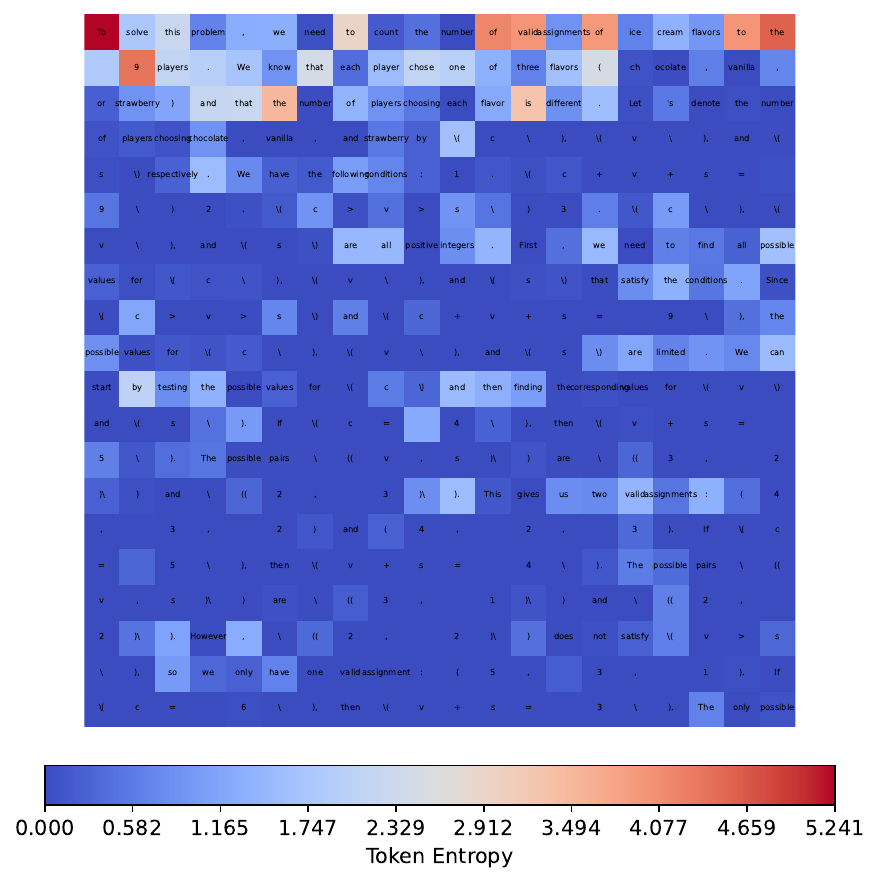}
	\caption{Random example 1 from AIME25 with \modelname{}}
	\label{fig:entropy-japo-1}
\end{figure*}

\begin{figure*}
	\includegraphics[width=\linewidth]{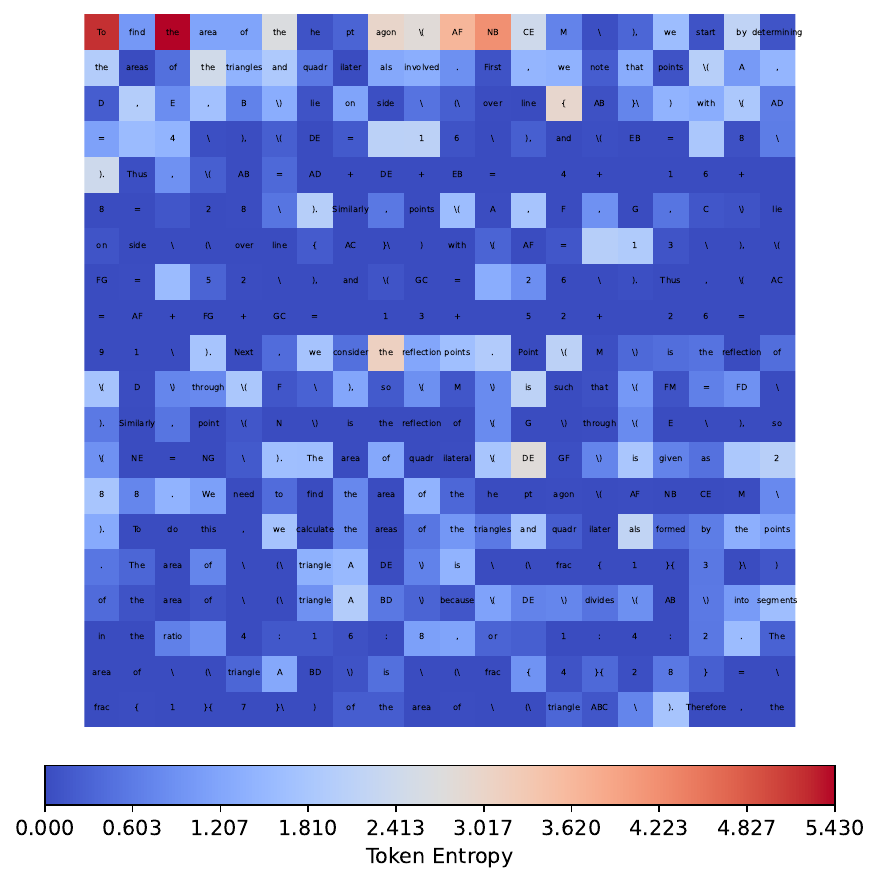}
	\caption{Random example 2 from AIME25 with GRPO}
	\label{fig:entropy-grpo-2}
\end{figure*}

\begin{figure*}
	\includegraphics[width=\linewidth]{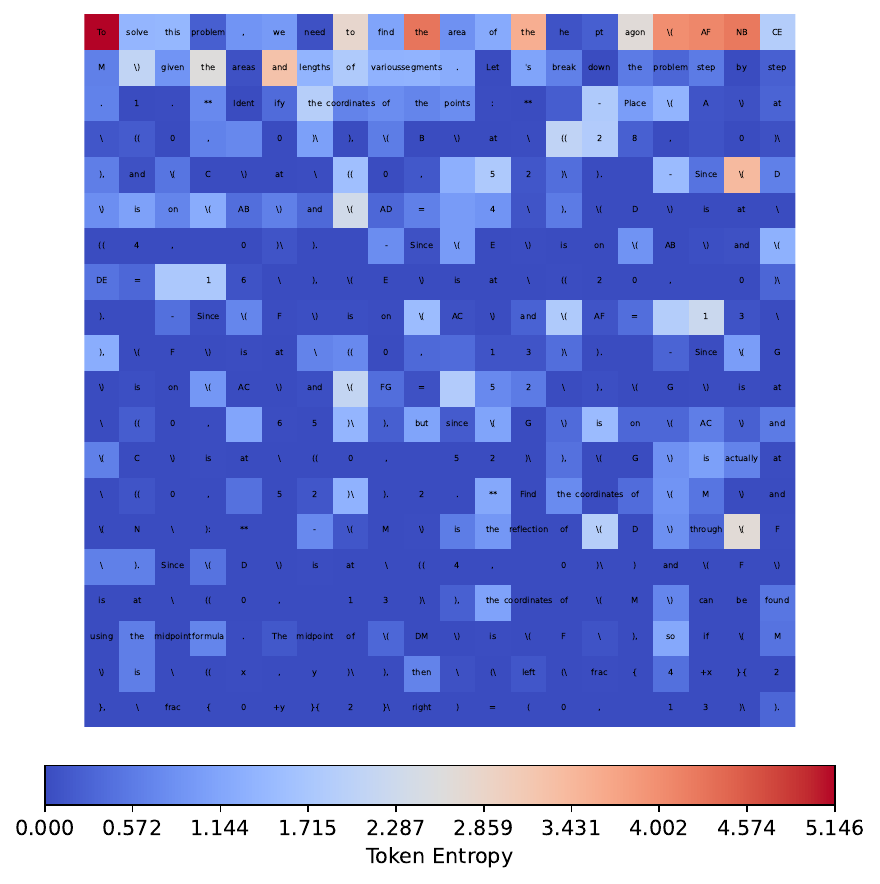}
	\caption{Random example 2 from AIME25 with \modelname{}}
	\label{fig:entropy-japo-2}
\end{figure*}

As shown in Figure~\ref{fig:entropy}, larger $n$ enables more aggressive token-level exploration in \modelname{}, leading to faster entropy reduction and earlier performance gains with minimal compute. In contrast, GRPO requires sampling $G$ full trajectories and converges more slowly despite higher computational cost. This highlights \modelname{}’s efficiency in refining reasoning behavior with fine-grained feedback. 
The entropy visualizations (Figures~\ref{fig:entropy-grpo-1} to \ref{fig:entropy-japo-2}) reveal that \modelname{}-aligned generations rely more heavily on low-entropy mathematical tokens (e.g., numbers, equations, variable names), while GRPO outputs include more high-entropy verbal phrases. This reflects a key difference in reasoning style and efficiency between the two methods.

\modelname{} promotes more confident and focused generation by emphasizing precise reasoning steps. Since mathematical tokens are drawn from a constrained, predictable space, they naturally have lower entropy. By training only on correct trajectories and applying token-level supervision selectively (via importance sampling), the proposed \modelname{} concentrates gradient signal on these informative, low-entropy tokens. This leads the model to internalize concise, symbol-driven reasoning patterns rather than verbose explanations.

This shift toward low-entropy math expressions has practical benefits: it reduces distractions, shortens reasoning paths, and increases the density of useful signal per token. As a result, \modelname{} produces more concise rationales and achieves better task performance.

\end{document}